\documentclass{article}
\PassOptionsToPackage{compress}{natbib}

\usepackage[table]{xcolor}
\usepackage{hhline}
\usepackage{xspace} 
\usepackage{todonotes}
\usepackage{algorithm}
\usepackage{booktabs}
\usepackage{graphicx} 
\usepackage{algpseudocode}
\usepackage{siunitx}
\usepackage{svg}
\usepackage{subcaption}
\usepackage{float}
\usepackage{inconsolata}
\usepackage{xcolor}
\usepackage{bbding}
\usepackage{amsfonts}
\usepackage{amsmath}
\usepackage{wrapfig}
\usepackage{enumitem}

\DeclareCaptionLabelFormat{andtable}{#1~#2  \&  \tablename~\thetable}
\usepackage[font=small,labelfont=bf,tableposition=top]{caption}

\definecolor{lightblue}{HTML}{e5eff5}
\definecolor{lightorange}{HTML}{ffecd7}
\definecolor{darkblue}{HTML}{0065a2}
\definecolor{darkorange}{HTML}{ffa23a}

\DeclareMathAlphabet\mathbfcal{OMS}{cmsy}{b}{n}

\usepackage[final]{corl_2023}

\newcommand{\algname}{VoxPoser\xspace}
\newcommand{\algabrvname}{VoxPoser\xspace}
\newcommand\blfootnote[1]{%
  \begingroup
  \renewcommand\thefootnote{}\footnote{#1}%
  \addtocounter{footnote}{-1}%
  \endgroup
}

\title{\algabrvname: Composable 3D Value Maps\\for Robotic Manipulation with Language Models}

\author{
  Wenlong Huang$^1$, Chen Wang$^1$, Ruohan Zhang$^1$, Yunzhu Li$^{1,2}$, Jiajun Wu$^1$, Li Fei-Fei$^1$\\
  $^1$Stanford University \quad $^2$University of Illinois Urbana-Champaign
}

\begin{document}
\maketitle

\begin{abstract}
    Large language models (LLMs) are shown to possess a wealth of actionable knowledge that can be extracted for robot manipulation in the form of reasoning and planning. Despite the progress, most still rely on pre-defined motion primitives to carry out the physical interactions with the environment, which remains a major bottleneck.
    In this work, we aim to synthesize robot trajectories, i.e., a dense sequence of 6-DoF end-effector waypoints, for a large variety of manipulation tasks given an \emph{open-set~of~instructions} and an \emph{open-set~of~objects}.
    We achieve this by first observing that LLMs excel at inferring affordances and constraints given a free-form language instruction. More importantly, by leveraging their code-writing capabilities, they can interact with a vision-language model (VLM) to compose 3D value maps to ground the knowledge into the observation space of the agent.
    The composed value maps are then used in a model-based planning framework to \emph{zero-shot} synthesize closed-loop robot trajectories with robustness to dynamic perturbations.
    We further demonstrate how the proposed framework can benefit from online experiences by efficiently learning a dynamics model for scenes that involve contact-rich interactions.
    We present a large-scale study of the proposed method in both simulated and real-robot environments, showcasing the ability to perform a large variety of everyday manipulation tasks specified in free-form natural language.
    Videos and code at \href{https://voxposer.github.io}{voxposer.github.io}.
    \blfootnote{Correspondence to Wenlong Huang \textless wenlongh@stanford.edu\textgreater.}
\end{abstract}


\begin{figure}[h]
    \centering
    \includegraphics[width=1.0\linewidth]{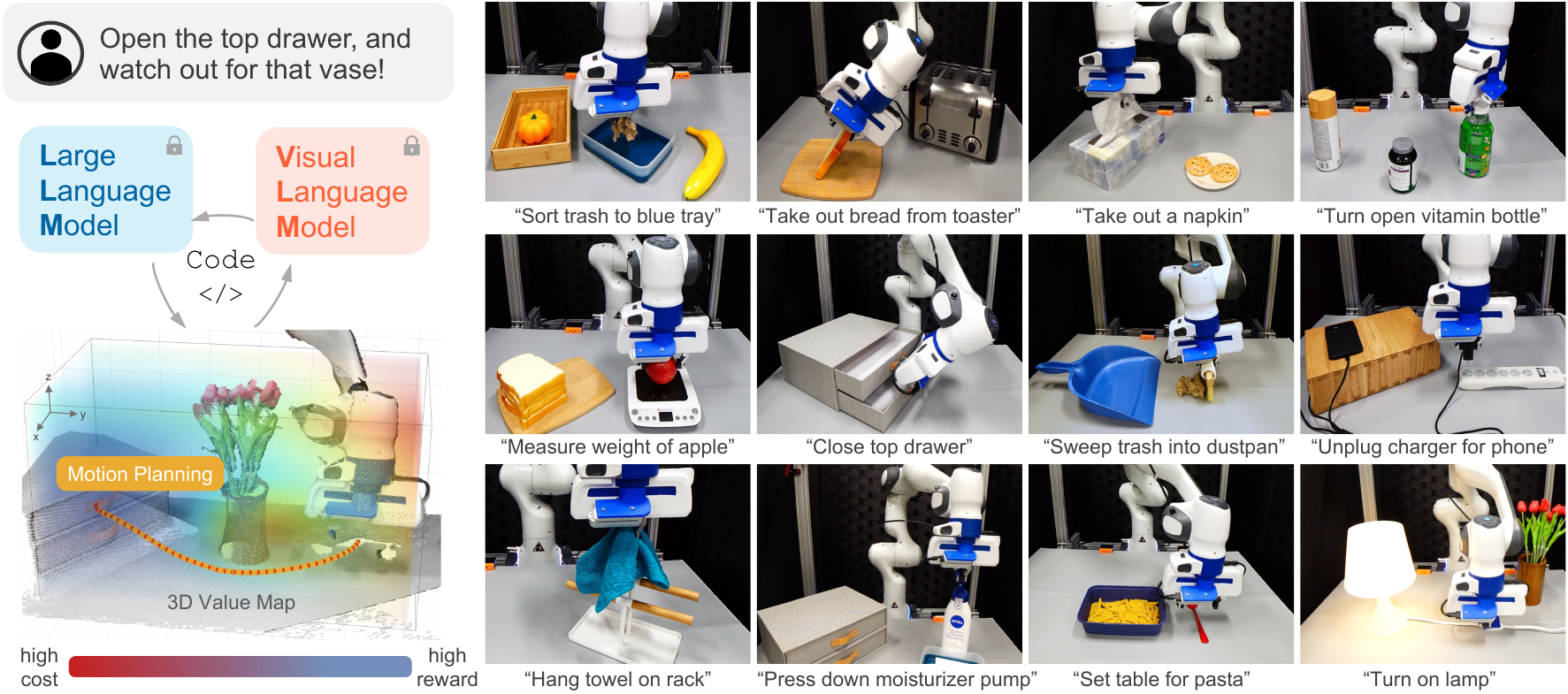}
    \vspace{-1.2em}
    \caption{\small \textbf{\textsc{\algname}} extracts language-conditioned \textbf{affordances} and \textbf{constraints} from LLMs and grounds them to the perceptual space using VLMs, using a code interface and without additional training to either component. The composed map is referred to as a 3D value map, which enables \textbf{zero-shot} synthesis of trajectories for large varieties of everyday manipulation tasks with an \textbf{open-set of instructions} and an \textbf{open-set of objects}.}
    \label{fig:teaser}
    \vspace{-1.55em}
\end{figure}

\section{Introduction}

Language is a compressed medium through which humans distill and communicate their knowledge and experience of the world. Large language models (LLMs) have emerged as a promising approach to capture this abstraction, learning to represent the world through projection into language space~\cite{brown2020language,openai2023gpt,chowdhery2022palm,bommasani2021opportunities}. While these models are believed to internalize generalizable knowledge as text, it remains a question about how to use it to enable embodied agents to physically \emph{act} in the real world.

We look at the problem of grounding abstract language instructions (e.g., ``set up the table'') in robot actions~\cite{tellex2020robots}. Prior works have leveraged lexical analysis to parse the instructions~\cite{tellex2011understanding,kollar2010toward,bollini2013interpreting}, while more recently language models have been used to decompose the instructions into a textual sequence of steps~\cite{huang2022language,ahn2022can,zeng2022socratic}.
However, to enable physical interactions with the environment, existing approaches typically rely on a repertoire of pre-defined motion primitives (i.e., skills) that may be invoked by an LLM or a planner, and this reliance on individual skill acquisition is often considered a major bottleneck of the system due to the lack of large-scale robotic data. The question then arises: how can we leverage the wealth of internalized knowledge of LLMs at the even fine-grained action level for robots, without requiring laborious data collection or manual designs for each individual primitive?

In addressing this challenge, we first note that it is impractical for LLMs to directly output control actions in text, which are typically driven by high-frequency control signals in high-dimensional space.
However, we find that LLMs excel at inferring language-conditioned \emph{affordances} and \emph{constraints}, and by leveraging their code-writing capabilities, they can compose dense 3D voxel maps that ground them in the visual space by orchestrating perception calls (e.g., via CLIP~\cite{radford2021learning} or open-vocabulary detectors~\citep{gu2021open,kamath2021mdetr,minderer2022simple}) and array operations (e.g., via NumPy~\cite{harris2020array}).
For example, given an instruction ``open the top drawer and watch out for the vase'', LLMs can be prompted to infer: 1) the top drawer handle should be grasped, 2) the handle needs to be translated outwards, and 3) the robot should stay away from the vase.
% While these are expressed in the text form, LLMs can generate Python code to invoke perception APIs to obtain spatial-geometric information of relevant objects or parts (e.g., ``handle'') and then manipulate the 3D voxels to prescribe reward or cost at relevant locations in observation space (e.g., the target location of the handle is assigned a high value while the surrounding of the vase is assigned low values).
By generating Python code to invoke perception APIs, LLMs can obtain spatial-geometric information of relevant objects or parts and then manipulate the 3D voxels to prescribe reward or cost at relevant locations in observation space (e.g., the handle region is assigned high values while the surrounding of the vase is assigned low values).
Finally, the composed value maps can serve as objective functions for motion planners to directly synthesize robot trajectories that achieve the given instruction
\footnote{The approach also bears resemblance and connections to potential field methods in path planning~\cite{hwang1992potential} and constrained optimization methods in manipulation planning~\cite{toussaint2022sequence}.}
, without requiring additional training data for each task or for the LLM. An illustration diagram and a subset of tasks we considered are shown in Fig.~\ref{fig:teaser}.

We term this approach \textsc{\textbf{\algname}}
, a formulation that extracts affordances and constraints from LLMs to compose 3D value maps in observation space for guiding robotic interactions.
% In particular, the method leverages LLMs to \emph{compose} key aspects for generating interaction trajectories rather than relying on robotic data that are often of limited amount or variability, effectively achieving generalization for \emph{open-set} instructions in a zero-shot manner.
% By integrating it into a model-based planning framework, we demonstrate closed-loop executions via model predictive control (MPC) that is robust to external disturbances.
Rather than relying on robotic data that are often of limited amount or variability, the method leverages LLMs for \emph{open-world reasoning} and VLMs for \emph{generalizable visual grounding} in a model-based planning framework that directly enables \emph{physical} robot actions.
We demonstrate its zero-shot generalization for \emph{open-set} instructions with \emph{open-set} objects for various everyday manipulation tasks.
We further showcase how~\algname can also benefit from limited online interactions to efficiently learn a dynamics model that involves contact-rich interactions.

\section{Related Works}
\label{sec:related}

\textbf{Grounding Language Instructions. }
Language grounding has been studied extensively both in terms of intelligent agents~\cite{andreas2017learning,zellers2021piglet,zellers2021merlot,shwartz2020unsupervised} and of robotics~\cite{winograd1971procedures,tellex2011understanding,blukis2020few,tellex2014asking,tellex2020robots,kollar2010toward,kollar2014grounding}, where language can be used as a tool for compositional goal specification~\cite{tellex2020robots,thomason2015learning,thomason2020jointly,jang2022bc,brohan2022rt,shah2022lm,cui2023no,stone2023open}, semantic anchor for training multi-modal representation~\cite{radford2021learning,nair2022r3m,ma2023liv}, or as an intermediate substrate for planning and reasoning~\cite{jansen2020visually,micheli2021language,sharma2021skill,huang2022language,ahn2022can,huang2022inner,li2023lampp}. Prior works have looked at using classical tools such as lexical analysis, formal logic, and graphical models to interpret language instructions~\cite{thomason2015learning,kollar2010toward,tellex2011understanding,kollar2014grounding}. More recently, end-to-end approaches, popularized by successful applications to offline domains~\cite{imagenet_cvpr09,krizhevsky2017imagenet,radford2019language,brown2020language}, have been applied to directly ground language instructions in robot interactions by learning from data with language annotations, spanning from model learning~\cite{nair2022learning}, imitation learning~\cite{shridhar2022cliport,shridhar2022peract,brohan2022rt,li2022pretrained,mees2022matters,mees2022grounding,sharma2022correcting,liu2022structformer,lynch2021language,lynch2022interactive,shao2021concept2robot}, to reinforcement learning~\cite{Luketina2019ASO,Andreas2017ModularMR,jiang2019language}.
Most closely related to our work is~\citet{sharma2022correcting}, where an end-to-end cost predictor is optimized via supervised learning to map language instructions to 2D costmaps, which are used to steer a motion planner to generate preferred trajectories in a collision-free manner. In contrast, we rely on pre-trained language models for their open-world knowledge and tackle the more challenging robotic manipulation in 3D.

\textbf{Language Models for Robotics. }
Leveraging pre-trained language models for embodied applications is an active area of research, where a large body of works focus on planning and reasoning with language models~\cite{huang2022language,ahn2022can,zeng2022socratic,chen2022open,shah2022lm,huang2022inner,singh2022progprompt,huang2022visual,raman2022planning,song2022llm,liu2023llm,vemprala2023chatgpt,lin2023text2motion,ding2023task,huang2023grounded,driess2023palm,yuan2023plan4mc,xie2023translating,lu2023multimodal,patel2023pretrained,jansen2020visually,wang2023voyager,yang2023pave}. To allow language models to perceive the physical environments, textual descriptions of the scene~\cite{huang2022inner,zeng2022socratic,singh2022progprompt} or perception APIs~\cite{liang2022code} can be given, vision can be used during decoding~\cite{huang2023grounded} or can be directly taken as input by multi-modal language models~\cite{driess2023palm,openai2023gpt}. In addition to perception, to truly bridge the perception-action loop, an embodied language model must also know how to \emph{act}, which typically is achieved by a library of pre-defined primitives.~\citet{liang2022code} showed that LLMs exhibit behavioral commonsense that can be useful for low-level control.
Despite the promising signs, hand-designed motion primitives are still required, and while LLMs are shown to be capable of composing \emph{sequential} policy logic, it remains unclear whether composition can happen at \emph{spatial} level.
A related line of works has also explored using LLMs for reward specification in the context of reward design~\cite{kwon2023reward}, exploration~\cite{tam2022semantic,mu2022improving,colas2020language,du2023guiding}, and preference learning~\cite{hu2023language}.
For robotic applications, concurrent works explored LLM-based reward generation~\cite{yu2023language,ha2023scaling,ma2023eureka,xie2023text2reward,zeng2023learning,rocamonde2023vision,perez2023larg}, among which~\citet{yu2023language} use MuJoCo~\cite{mujoco} as a high-fidelity physics model for model predictive control.
In contrast, we focus exclusively on grounding the reward generated by LLMs in the \emph{3D observation space} of the robot.

% Concurrent work also explores leveraging foundation models in imitation learning for pick-and-place robotic tasks~\cite{yang2023pave}.

\textbf{Learning-based Trajectory Optimization. }
Many works have explored leveraging learning-based approaches for trajectory optimization.
While the literature is vast, they can be broadly categorized into those that learn the models~\cite{lenz2015deepmpc,hewing2020learning,chang2016compositional,battaglia2016interaction,xu2019densephysnet,byravan2017se3,nagabandi2020deep,sanchez2018graph,li2018learning} and those that learn the cost/reward or constraints~\cite{finn2016guided,fu2017learning,driess2020deep,amos2018differentiable,sharma2022correcting,mittal2022articulated}, where data are typically collected from in-domain interactions.
To enable generalization in the wild, a parallel line of works has explored learning task specification from large-scale offline data~\cite{bahl2022human,bahl2023affordances,wang2023mimicplay,ma2023liv,nair2022r3m,nair2022learning,bharadhwaj2023zero,ma2022vip,shao2021concept2robot}, particularly egocentric videos~\cite{damen2018scaling,grauman2022ego4d}, or leveraging pre-trained foundation models~\cite{cui2022can,yu2023scaling,mandi2022cacti,stone2023open,xiao2022robotic,wang2022generalizable}.
The learned cost functions are then used by reinforcement learning~\cite{cui2022can,ma2022vip,li2023behavior}, imitation learning~\cite{wang2023mimicplay,bahl2023affordances}, or trajectory optimization~\cite{bahl2022human,ma2023liv} to generate robot actions.
In this work, we leverage LLMs for \emph{zero-shot} \emph{in-the-wild} cost specification with superior generalization.
Compared to prior works that leverage foundation models, we ground the cost directly in \emph{3D observation space} with \emph{real-time} visual feedback, which makes~\algname amenable to closed-loop MPC that's robust in execution.

\section{Method}\label{sec:method}

We first provide the formulation of~\algname as an optimization problem (Sec.~\ref{sec:formulation}). Then we describe how~\algname can be used as a general zero-shot framework to map language instructions to 3D value maps (Sec.~\ref{sec:voxposer}). We subsequently demonstrate how trajectories can be synthesized in closed-loop for robotic manipulation (Sec.~\ref{sec:mpc}). While zero-shot in nature, we demonstrate how~\algname can learn from online interactions to efficiently solve contact-rich tasks (Sec.~\ref{sec:dynamics}).

\subsection{Problem Formulation}\label{sec:formulation}

Consider a manipulation problem given as a \emph{free-form} language instruction $\mathcal{L}$ (e.g., ``open the top drawer''). Generating robot trajectories according to $\mathcal{L}$ can be very challenging because $\mathcal{L}$ may be arbitrarily long-horizon or under-specified (i.e., requires contextual understanding). Instead, we focus on individual phases (sub-tasks) of the problem $\ell_i$ that distinctively specify a manipulation task (e.g., ``grasp the drawer handle'', ``pull open the drawer''), where the decomposition $\mathcal{L} \to (\ell_1, \ell_2, \dots, \ell_n)$ is given by a high-level planner (e.g., an LLM or a search-based planner)
\footnote{Note that the decomposition and sequencing of these sub-tasks are also done by LLMs in this work, though we do not investigate this aspect extensively as it is not the focus of our contributions.}.
The central problem investigated in this work is to generate a motion trajectory $\tau_i^{\mathbf{r}}$ for robot $\mathbf{r}$ and each manipulation phase described by instruction $\ell_i$. We represent $\tau_i^{\mathbf{r}}$ as a sequence of dense end-effector waypoints to be executed by an Operational Space Controller~\cite{khatib1987unified}, where each waypoint consists of a desired 6-DoF end-effector pose, end-effector velocity, and gripper action. However, it is worth noting that other representations of trajectories, such as joint space trajectories, can also be used. Given the $i$-th sub-task described by $\ell_i$, we formulate an optimization problem defined as follows:

\vspace{-1.2em}
\begin{equation}\label{eq:formulation}
\min_{\tau_i^{\mathbf{r}}} \left\{ \mathcal{F}_{task}(\mathbf{T}_i, \ell_i) + \mathcal{F}_{control}(\tau_i^{\mathbf{r}}) \right\} \quad \text{subject to} \quad \mathcal{C}(\mathbf{T}_i)
\end{equation}
\vspace{-1.2em}

where $\mathbf{T}_i$ is the evolution of environment state, and $\tau_i^{\mathbf{r}} \subseteq \mathbf{T}_i$ is the robot trajectory. $\mathcal{F}_{task}$ scores the extent of $\mathbf{T}_i$ completes the instruction $\ell_i$ while $\mathcal{F}_{control}$ specifies the control costs, e.g., to encourage $\tau_i^{\mathbf{r}}$ to minimize total control effort or total time. 
$\mathcal{C}(\mathbf{T}_i)$ denotes the dynamics and kinematics constraints, which are enforced by the known model of the robot and a physics-based or learning-based model of the environment.
% The initial condition constraint is represented by $\mathbf{T}_i(0) = \mathbf{x}$, where $\mathbf{x}$ is the initial state of the environment at the start of sub-task $\ell_i$.
By solving this optimization for each sub-task $\ell_i$, we obtain a sequence of robot trajectories that collectively achieve the overall task specified by the instruction $\mathcal{L}$.

\begin{figure}
    \centering
    \includegraphics[width=1.0\linewidth]{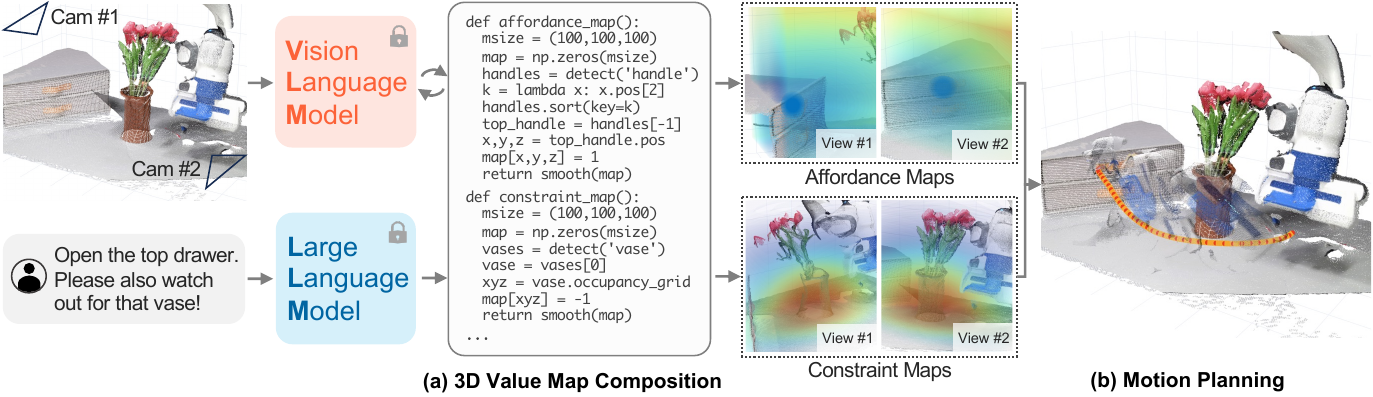}
    \vspace{-1.2em}
    \caption{\small \textbf{Overview of~\textsc{\algname}}. Given the RGB-D observation of the environment and a language instruction, LLMs generate code, which interacts with VLMs, to produce a sequence of 3D affordance maps and constraint maps (collectively referred to as value maps) grounded in the observation space of the robot \textbf{(a)}. The composed value maps then serve as objective functions for motion planners to synthesize trajectories for robot manipulation \textbf{(b)}. The entire process does not involve any additional training.}
    \label{fig:method}
    \vspace{-1.7em}
\end{figure}

\vspace{-.5em}
\subsection{Grounding Language Instruction via \algabrvname}\label{sec:voxposer}

Calculating $\mathcal{F}_{task}$ with respect to free-form language instructions is extremely challenging, not only because of the rich space of semantics language can convey but also because of the lack of robot data labeled with $\mathbf{T}$ and $\ell$. However, we provide a critical observation that a large number of tasks can be characterized by a voxel value map
$\mathbf{V} \in \mathbb{R}^{w \times h \times d}$
in robot's observation space, which guides the motion of an ``entity of interest'' in the scene, such as the robot end-effector, an object, or an object part.
For example, consider the task ``open the top drawer'' and its first sub-task ``grasp the top drawer handle'' (inferred by LLMs) in Fig.~\ref{fig:method}. The ``entity of interest'' is the robot end-effector, and the voxel value map should reflect the attraction toward the drawer handle. By further commanding ``watch out for the vase'', the map can also be updated to reflect the repulsion from the vase.
We denote the ``entity of interest'' as $\mathbf{e}$ and its trajectory as $\tau^{\mathbf{e}}$.
Using this voxel value map for a given instruction $\ell_i$, $\mathcal{F}_{task}$ can be approximated by accumulating the values of $\mathbf{e}$ traversing through $\mathbf{V}_i$, formally calculated as $\mathcal{F}_{task} = -\sum_{j=1}^{|\tau_i^{\mathbf{e}}|} \mathbf{V}(p^{\mathbf{e}}_{j})$, where $p^{\mathbf{e}}_{j} \in \mathbb{N}^3$ is the discretized $(x,y,z)$ position of $\mathbf{e}$ at step $j$.

Notably, we observe large language models, by being pre-trained on Internet-scale data, exhibit capabilities not only to identify the ``entity of interest'' but also to compose value maps that accurately reflect the task instruction by writing Python programs. Specifically, when an instruction is given as a comment in the code, LLMs can be prompted to 1)~call perception APIs (which invoke vision-language models (VLM) such as an open-vocabulary detector~\citep{gu2021open,kamath2021mdetr,minderer2022simple}) to obtain spatial-geometrical information of relevant objects, 2)~generate NumPy operations to manipulate 3D arrays, and 3)~prescribe precise values at relevant locations.
We term this approach as~\textbf{\textsc{\algname}}. Concretely, we aim to obtain a voxel value map $\mathbf{V}_i^t = \text{\algname}(\mathbf{o}^t, \ell_i)$ by prompting an LLM and executing the code via a Python interpreter, where $\mathbf{o}^t$ is the RGB-D observation  at time $t$ and $\ell_i$ is the current instruction.
Additionally, because $\mathbf{V}$ is often sparse, we densify the voxel maps via smoothing operations, as they encourage smoother trajectories optimized by motion planners.

\begin{figure}[t]
    \centering
    \includegraphics[width=0.95\linewidth]{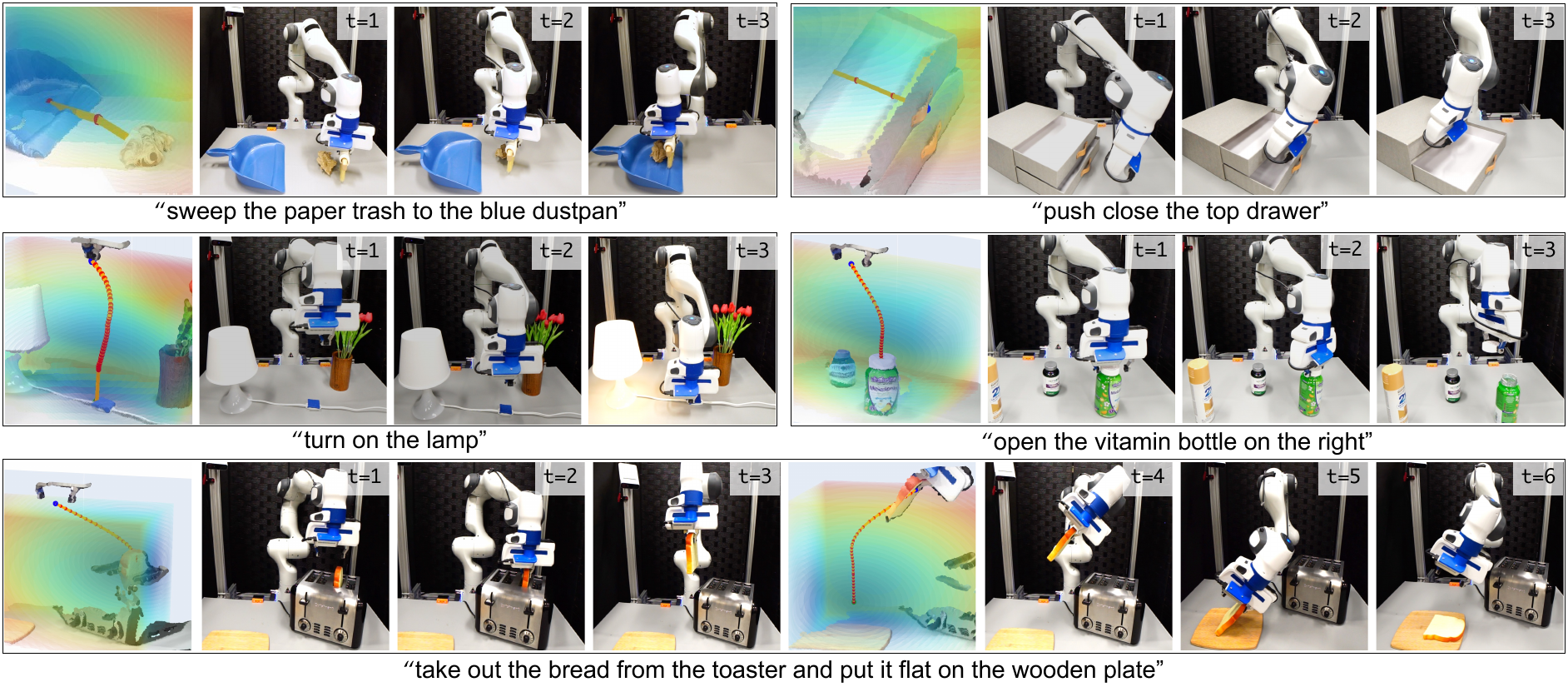}
    \vspace{-0.5em}
    \caption{\small Visualization of composed 3D value maps and rollouts in real-world environments. The top row demonstrates where ``entity of interest'' is an object or part, and the value maps guide them toward target positions. The bottom two rows showcase tasks where ``entity of interest'' is the robot end-effector. The bottom-most task involves two phases, which are also orchestrated by LLMs.}
    \label{fig:franka}
    \vspace{-1.75em}
\end{figure}

\textbf{Additional Trajectory Parametrization. }
The above formulation of~\algname uses LLMs to compose $\mathbf{V}: \mathbb{N}^3 \to \mathbb{R}$ to map from discretized coordinates in voxel space to a real-valued ``cost'', which we can use to optimize a path consisting only of the positional terms.
To extend to SE$(3)$ poses, we can also use LLMs to compose rotation maps $\mathbf{V}_r: \mathbb{N}^3 \to \text{SO}(3)$ at coordinates relevant to the task objectives (e.g., ``end-effector should face the support normal of the handle''). Similarly, we further compose gripper maps $\mathbf{V}_g: \mathbb{N}^3 \to \{0, 1\}$ to control gripper open/close and velocity maps $\mathbf{V}_v: \mathbb{N}^3 \to \mathbb{R}$ to specify target velocities. Note that while these additional trajectory parametrizations are not mapped to a real-valued ``cost'', they can also be factored in the optimization procedure (Equation~\ref{eq:formulation}) to parametrize the trajectories.

\subsection{Zero-Shot Trajectory Synthesis with~\algname}\label{sec:mpc}
After obtaining the task cost $\mathcal{F}_{task}$, we can now approach the full problem defined in Equation~\ref{eq:formulation} to plan a motion trajectory.
We use simple zeroth-order optimization by randomly sampling trajectories and scoring them with the proposed objective.
The optimization is implemented in a model predictive control framework that iteratively replans the trajectory at every step using the current observation to robustly execute the trajectories even under dynamic disturbances
\footnote{Although involving an LLM in the loop, closed-loop execution is possible because the generated code remains the same throughout task $\ell_i$, which allows us to cache its output for the current task.}
, where either a learned or physics-based model can be used. However, because~\algname effectively provides ``dense rewards'' in the observation space and we are able to replan at every step, we surprisingly find that the overall system can already achieve a large variety of manipulation tasks considered in this work even with simple heuristics-based models.
Since some value maps are defined over ``entity of interest'', which may not necessarily be the robot, we also use the dynamics model to find the needed robot trajectory to minimize the task cost (i.e., what interactions between the robot and the environment achieve the desired object motions).
% Because our method is agnostic to specific instantiations of motion planning, we leave extended discussion of our implementation to Sec.~\ref{sec:exp}.

\subsection{Efficient Dynamics Learning with Online Experiences}\label{sec:dynamics}
While Sec.~\ref{sec:mpc} presents a zero-shot framework for synthesizing trajectories for robot manipulation,~\algname can also benefit from online experiences by efficiently learning a dynamics model.
Consider the standard setup where a robot interleaves between 1) collecting environment transition data $(\mathbf{o}_t, \mathbf{a}_t,  \mathbf{o}_{t+1})$, where $\mathbf{o}_t$ is the environment observation at time $t$ and $\mathbf{a}_t = \text{MPC}(\mathbf{o}_t)$, and 2) training a dynamics model $\mathbf{g}_{\theta}$ parametrized by $\theta$ by minimizing the L2 loss between predicted next observation $\hat{\mathbf{o}}_{t+1}$ and $\mathbf{o}_{t+1}$. A critical component that determines the learning efficiency is the action sampling distribution $P(\mathbf{a}_t | \mathbf{o}_t)$ in MPC, which typically is a random distribution over the full action space $\mathbf{A}$. This is often inefficient when the goal is to solve a particular task, such as opening a door, because most actions do not interact with the relevant objects in the scene (i.e., the door handle) nor do they necessarily interact with the objects in a meaningful way (i.e., pressing down the door handle). Since~\algname synthesizes robot trajectories with LLMs, which have a wealth of commonsense knowledge, the zero-shot synthesized trajectory $\tau^{\mathbf{r}}_0$ can serve as a useful prior to bias the action sampling distribution $P(\mathbf{a}_t | \mathbf{o}_t, \tau^{\mathbf{r}}_0)$, which can significantly speed up the learning process.
In practice, this can be implemented by only sampling actions in the vicinity of $\tau^{\mathbf{r}}_0$ by adding small noise $\varepsilon$ to encourage local exploration instead of exploring in the full action space $\mathbf{A}$.

\section{Experiments and Analysis}\label{sec:exp}

% We first discuss our design choices for implementing~\algname below. Then we validate directly in real-world systems whether~\algname can perform everyday manipulation tasks in Sec.~\ref{sec:real}. We also present a detailed quantitative study of the generalization performance of~\algname compared to learned and LLM-based baselines in simulation in Sec.~\ref{sec:sim}. We further demonstrate how~\algname can benefit from only limited online experience to learn a dynamics model for contact-rich tasks in Sec.~\ref{sec:dynamics-exp}. Finally, we study the source of errors in the overall systems and discuss how improvement can be made in Sec.~\ref{sec:error}.
We first discuss our implementation details. Then we validate~\algname for real-world everyday manipulation (Sec.~\ref{sec:real}). We also study its generalization in simulation (Sec.~\ref{sec:sim}). We further demonstrate how~\algname enables efficient learning of more challenging tasks (Sec.~\ref{sec:dynamics-exp}). Finally, we analyze its source of errors and discuss how improvement can be made (Sec.~\ref{sec:error}). 

% \subsection{Implementation of~\algname}\label{sec:implementation}

% Herein we discuss our instantiation of~\algname. We focus our discussions on design choices shared between the simulated and real-world domains. More details about the environment setup in each domain can be found in Appendix.

\textbf{LLMs and Prompting.}
% We follow prompting structure by~\citet{liang2022code}, which recursively calls LLMs using their own generated code, where each language model program (LMP) is responsible for a unique functionality (e.g., processing perception calls). We use GPT-4~\cite{openai2023gpt} from~\href{https://openai.com/api/}{OpenAI API}. Prompts are in Appendix.
We follow prompting structure by~\citet{liang2022code}, which recursively calls LLMs using their own generated code, where each language model program (LMP) is responsible for a unique functionality (e.g., processing perception calls). We use GPT-4~\cite{openai2023gpt} from~\href{https://openai.com/api/}{OpenAI API}.
For each LMP, we include 5-20 example queries and corresponding responses as part of the prompt. An example can be found in Fig.~\ref{fig:method} (simplified for clarity). Full prompts are in Appendix.

\textbf{VLMs and Perception.}
Given an object/part query from LLMs, we first invoke open-vocab detector OWL-ViT~\cite{minderer2022simple} to obtain a bounding box, then feed it into Segment Anything~\cite{kirillov2023segment} to obtain a mask, and finally track the mask using video tracker XMEM~\cite{cheng2022xmem}. The tracked mask is used with RGB-D observation to reconstruct the object/part point cloud.

\textbf{Value Map Composition.}
We define the following types of value maps: affordance, avoidance, end-effector velocity, end-effector rotation, and gripper action. Each type uses a different LMP, which takes in an instruction and outputs a voxel map of shape $(100, 100, 100, k)$, where $k$ differs for each value map (e.g., $k=1$ for affordance and avoidance as it specifies cost, and $k=4$ for rotation as it specifies SO$(3)$). We apply Euclidean distance transform to affordance maps and Gaussian filters for avoidance maps. On top of value map LMPs, we define two high-level LMPs to orchestrate their behaviors: \texttt{planner} takes user instruction $\mathcal{L}$ as input (e.g., ``open drawer'') and outputs a sequence of sub-tasks $\ell_{1:N}$, and \texttt{composer} takes in sub-task $\ell_i$ and invokes relevant value map LMPs with detailed language parameterization.

\textbf{Motion Planner. }
We consider only affordance and avoidance maps in the planner optimization, which finds a sequence of collision-free end-effector positions $p_{1:N} \in \mathbb{R}^3$ using greedy search. Then we enforce other parametrization at each $p$ by the remaining value maps (e.g., rotation map, velocity map). The cost map used by the motion planner is computed as the negative of the weighted sum of normalized affordance and avoidance maps with weights $2$ and $1$. After a 6-DoF trajectory is synthesized, the first waypoint is executed, and then a new trajectory is re-planned at 5 Hz.

% \textbf{Environment Dynamics Model. }
% For tasks in which the specified ``entity of interest'' is the robot, we assume identity environment dynamics while replanning at every step to account for the latest observation. For tasks in which the ``entity of interest'' is an object, we study only a planar pushing model parametrized by contact point, push direction, and push distance. The heuristic-based dynamics model translates an input point cloud along the push direction by the push distance. We use MPC with random shooting to optimize for the action parameters. Then a pre-defined pushing primitive is executed based on the action parameters. However, we note that a primitive is not necessary when action parameters are defined over the end-effector or joint space of the robot, which would likely yield smoother trajectories but takes more time for optimization.
\textbf{Dynamics Model. }
We use the known robot dynamics model in all tasks, where it is used in motion planning for the end-effector to follow the waypoints. For the majority of our considered tasks where the ``entity of interest'' is the robot, no environment dynamics model is used (i.e., scene is assumed to be static), but we replan at every step to account for the latest observation.
For tasks in which the ``entity of interest'' is an object, we study only a planar pushing model parametrized by contact point, push direction, and push distance. We use a heuristic-based dynamics model that translates an input point cloud along the push direction by the push distance.
We use MPC with random shooting to optimize for the action parameters. Then a pre-defined pushing primitive is executed based on the action parameters. However, we note that a primitive is not necessary when action parameters are defined over the end-effector or joint space of the robot, which would likely yield smoother trajectories but takes more time for optimization.
We also explore the use of a learning-based dynamics model in Section~\ref{sec:dynamics-exp}, which enables VoxPoser to benefit from online experiences.

\begin{figure}
\begin{minipage}{.45\textwidth}
    \centering
    \footnotesize
    \setlength\tabcolsep{2pt}
    \renewcommand{\arraystretch}{0.95}
    \begin{tabular}{@{}cccccc@{}}
        \toprule
         & \multicolumn{2}{c}{\textbf{LLM + Prim.~\cite{liang2022code}}} & \multicolumn{2}{c}{\textbf{\algabrvname}}  \\
        \cmidrule(lr){2-3} \cmidrule(lr){4-5}
        \textbf{Task} & Static & Dist. & Static & Dist. \\
        \midrule
        Move \& Avoid & 0/10 & 0/10 & 9/10 & 8/10 \\
        Set Up Table & 7/10 & 0/10 & 9/10 & 7/10 \\
        Close Drawer & 0/10 & 0/10 & 10/10 & 7/10 \\ 
        Open Bottle & 5/10 & 0/10 & 7/10 & 5/10 \\
        Sweep Trash & 0/10 & 0/10 & 9/10 & 8/10 \\
        \midrule
        \textbf{Total} & 24.0\% & 0.0\% & 88.0\% & 70.0\%\\
        \bottomrule
    \end{tabular}
\vspace{.5em}
\captionof{table}{\small Success rate in real-world domain.~\algname performs everyday manipulation tasks with high success and is more robust to disturbances than the baseline using action primitives.}
\label{tab:real}
\end{minipage}%
\hspace{2mm}
\begin{minipage}{.52\textwidth}
    \centering
    \footnotesize
    \setlength\tabcolsep{1.3pt}
    \renewcommand{\arraystretch}{0.95}
    \vspace{-0.5em}
    \begin{tabular}{@{}ccccc@{}}
        \toprule
         & & \multicolumn{1}{c}{\textbf{U-Net}} & \multicolumn{2}{c}{\textbf{Language Models}}  \\
        \cmidrule(lr){3-3} \cmidrule(lr){4-5}
        \textbf{Train/Test} & \textbf{Category} & \textbf{MP~\cite{sharma2022correcting}} & \textbf{Prim.~\cite{liang2022code}}  & \textbf{MP (Ours)}  \\
        \midrule
        SI SA & Object Int. & 21.0\% & 41.0\% & 64.0\%  \\
        SI SA & Composition & 53.8\% & 43.8\% & 77.5\%  \\
        \midrule
        SI UA & Object Int. & 3.0\% & 46.0\% & 60.0\%  \\
        SI UA & Composition & 3.8\% & 25.0\% & 58.8\%  \\
        \midrule
        UI UA & Object Int. & 0.0\% & 17.5\% & 65.0\%  \\
        UI UA & Composition & 0.0\% & 25.0\% & 76.7\%  \\
        \bottomrule
    \end{tabular}
\vspace{.5em}
\captionof{table}{\small Success rate in simulated domain. ``SI'' and ``UI'' are seen and unseen instructions. ``SA'' and ``UA'' are seen and unseen attributes.
~\algname outperforms both baselines across 13 tasks from two categories on both seen and unseen tasks and maintains similar success rates.}\label{tab:sim}
\end{minipage}
\vspace{-2.8em}
\end{figure}

\subsection{\algname for Everyday Manipulation Tasks}\label{sec:real}

We study whether~\algname can zero-shot synthesize robot trajectories to perform everyday manipulation tasks in the real world.
% We set up a real-world tabletop environment with a Franka Emika Panda robot. More details can be found in Appendix~\ref{app:real}.
Details of the environment setup can be found in Appendix~\ref{app:real}.
While the proposed method can generalize to an open-set of instructions and an open-set of objects as shown in Fig.~\ref{fig:teaser}, we pick 5 representative tasks to provide quantitative evaluations in Table~\ref{tab:real}.
Qualitative results including environment rollouts and value map visualizations are shown in Fig.~\ref{fig:franka}.
We find that~\algname can effectively synthesize robot trajectories for everyday manipulation tasks with a high average success rate. Due to fast replanning capabilities, it is also robust to external disturbances, such as moving targets/obstacles and pulling the drawer open after it has been closed by the robot.
% Particularly, by leveraging rich world knowledge in LLMs, we can extract language-conditioned affordances for diverse scenes and objects.
% For example, LLMs can infer that a bottle can be opened by turning counter-clockwise around the $z$-axis. Subsequently,~\algname can ground this knowledge in the observation space that directly guides motion planners to complete the tasks.
We further compare to a variant of Code as Policies~\cite{liang2022code} that uses LLMs to parameterize a pre-defined list of simple primitives (e.g., \texttt{move\_to\_pose}, \texttt{open\_gripper}). We find that compared to chaining sequential policy logic, the ability to \emph{compose spatially} while considering other constraints under a joint optimization scheme is a more flexible formulation, unlocking the possibility for more manipulation tasks and leading to more robust execution.
% In particular, by leveraging the composed spatial maps in MPC,~\algname can effectively recover from external disturbances, such as moving targets/obstacles and pulling the drawer open after it has been closed by the robot.

\subsection{Generalization to Unseen Instructions and Attributes}\label{sec:sim}
% Herein we investigate the generalization capabilities of~\algname. To provide rigorous quantitative results, we set up a simulated environment that mirrors our real-world setup~\cite{xiang2020sapien}, but with a fixed list of objects (consisting of a cabinet along with 10 colored blocks and lines) and a fixed list of 13 templated instructions (e.g., ``push [obj] to [pos]''), where [obj] and [pos] are attributes that are randomized over a pre-defined list.
% Herein we investigate the generalization capabilities of~\algname.
To provide rigorous quantitative evaluations on generalization, we set up a simulated block-world environment that mirrors our real-world robot setup~\cite{xiang2020sapien,mo2021where2act} but features 13 highly-randomizable tasks with 2766 unique instructions.
Eash task comes with a templated instruction (e.g., ``push [obj] to [pos]'') that contains randomizable attributes chosen from a pre-defined list. Details are in Appendix~\ref{app:sim}.
Seen instructions/attributes may appear in the prompt (or in the training data for supervised baselines).
% The instructions and attributes are divided into seen and unseen groups, where seen instructions/attributes may appear in the prompt (or in the training data for supervised baselines).
The tasks are grouped into 2 categories, where ``Object Interactions'' are tasks that require interactions with objects, and ``Spatial Composition'' are tasks involving spatial constraints (e.g., moving slower near a particular object).
For baselines, we ablate the two components of~\algname, LLM and motion planner, by comparing to a variant of~\cite{liang2022code} that combines an LLM with primitives and to a variant of~\cite{sharma2022correcting} that learns a U-Net~\cite{ronneberger2015u} to synthesize costmaps for motion planning. Table~\ref{tab:sim} shows the success rates averaged across 20 episodes per task.
% \algname outperforms both baselines in both categories, especially on unseen instructions or attributes.
We find~\algname exhibits superior generalization in all scenarios.
Compared to learned cost specification, LLMs generalize better by explicitly reasoning about affordances and constraints. On the other hand, grounding LLM knowledge in robot perception through \emph{value map composition} rather than directly specifying primitive parameters offers more flexibility and better generalization.

\subsection{Efficient Dynamics Learning with Online Experiences}\label{sec:dynamics-exp}

\begin{figure}[t]
\begin{minipage}{.54\textwidth}
    \centering
    \footnotesize
    \setlength\tabcolsep{0.4pt}
    \renewcommand{\arraystretch}{1.05}
    \begin{tabular}{@{}ccccccc@{}}
\toprule
 & \multicolumn{1}{c}{\textcolor{gray}{\textbf{Zero-Shot}}} & \multicolumn{2}{c}{\textbf{No Prior}} & \multicolumn{2}{c}{\textbf{w/ Prior}} \\
\cmidrule(lr){2-2} \cmidrule(lr){3-4} \cmidrule(lr){5-6}
\textbf{Task} & \textcolor{gray}{Success} & Success & Time(s) & Success & Time(s) \\
\midrule
Door & \textcolor{gray}{6.7\%\tinyy{$\pm$4.4\%}} & 58.3\tinyy{$\pm$4.4\%}& TLE & 88.3\%\tinyy{$\pm$1.67\%} & 142.3\tinyy{$\pm$22.4} \\
Window & \textcolor{gray}{3.3\%\tinyy{$\pm$3.3\%}} & 36.7\%\tinyy{$\pm$1.7\%} & TLE & 80.0\%\tinyy{$\pm$2.9\%} & 137.0\tinyy{$\pm$7.5} \\
Fridge & \textcolor{gray}{18.3\%\tinyy{$\pm$3.3\%}} & 70.0\%\tinyy{$\pm$2.9\%} & TLE & 91.7\%\tinyy{$\pm$4.4\%} & 71.0\tinyy{$\pm$4.4} \\
\bottomrule
\end{tabular}
\vspace{.5em}
% \captionof{table}{Zero-shot synthesized trajectories by~\algname, though limiting at contact-rich tasks, can serve as priors for efficient learning from online interactions. Compared to exploring without the priors,~\algname can learn to open various articulated objects with complex structures in under 3 minutes. TLE (time limit exceeded) means exceeding 12 hours. Results are reported over 3 runs different seeds.}
\captionof{table}{\algname enables efficient dynamics learning by using zero-shot synthesized trajectories as prior. TLE (time limit exceeded) means exceeding 12 hours. Results are reported over 3 runs different seeds.}
\label{tab:dynamics}
\end{minipage}%
\hspace{1.2mm}
\begin{minipage}{.455\textwidth}
    \centering
    \footnotesize
    \setlength\tabcolsep{1.3pt}
    \renewcommand{\arraystretch}{1.05}
    \vspace{-0.5em}
    \includegraphics[width=0.95\linewidth]{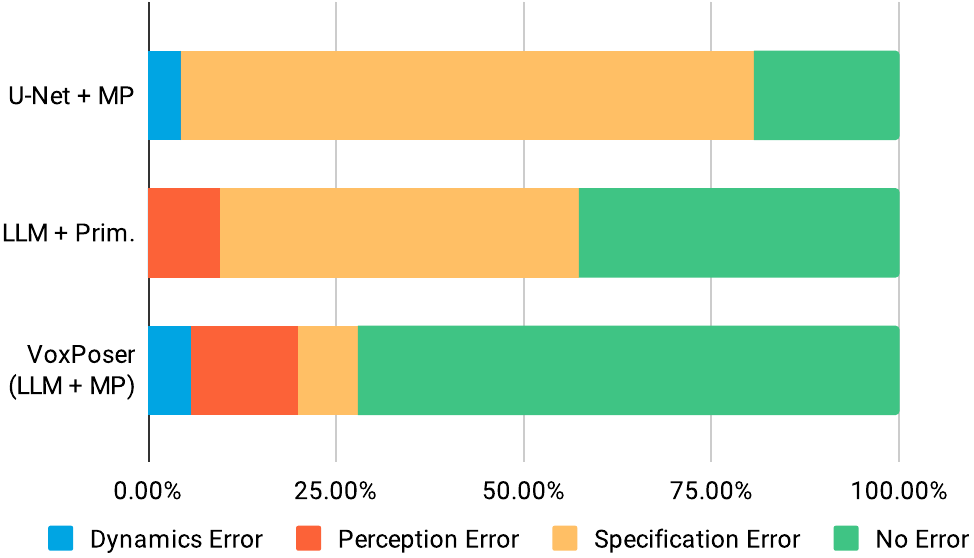}
    \caption{\small Error breakdown of components.~\algname significantly reduces specification error.}
    \label{fig:error}
\end{minipage}
\vspace{-2.0em}
\end{figure}

% Despite having zero-shot generalization to unseen instructions, we investigate how~\algname can also benefit from online interactions in tasks that involve more challenging contact-rich dynamics, as many behavioral nuances may not be present in the LLMs.
As discussed in Sec.~\ref{sec:dynamics}, we investigate how~\algname can optionally benefit from online experiences for tasks that involve more intricacies of contact, such as opening doors, fridges, and windows, in a simulated environment.
% To this end, we investigate a suite of simulated tasks involving interactions with common articulated objects, including opening doors, fridges, and windows.
% We hypothesize that while these are challenging tasks for autonomous agents due to difficult exploration, trajectories zero-shot synthesized by~\algname would provide useful hints for exploration (e.g., ``handle needs to be pressed down first in order to open a door'').
Specifically,  we first synthesize $k$ zero-shot trajectories using~\algname, each represented as a sequence of end-effector waypoints, that act as priors for exploration (e.g., ``handle needs to be pressed down first in order to open a door'').
Then an MLP dynamics model is learned through an iterative procedure where the agent alternates between data collection and model learning.
% The initial synthesized trajectories are used as prior in the action sampling distribution of MPC, where we add $\varepsilon \sim \mathcal{N}(0, \sigma^2)$ to each waypoint in $\tau^{\mathbf{r}}_0$ to encourage local exploration.
During data collection, we add $\varepsilon \sim \mathcal{N}(0, \sigma^2)$ to each waypoint in $\tau^{\mathbf{r}}_0$ to encourage local exploration.
% Results are shown in Table~\ref{tab:dynamics}.
% For these tasks involving complex interactions with articulated objects, we find trajectories zero-shot synthesized by~\algname are meaningful but insufficient.
As shown in Tab.~\ref{tab:dynamics}, we find zero-shot synthesized trajectories are typically meaningful but insufficient.
However, we can learn an effective dynamics model with less than 3 minutes of online interactions by using these trajectories as exploration prior, leading to high eventual success rates.
% In comparison, without the prior, it is extremely difficult to learn a dynamics model because most actions do not lead to meaningful environment changes. In all cases, the experiments exceed the maximum 12-hour limit.
In comparison, exploring without prior all exceed the maximum 12-hour limit.

\subsection{Error Breakdown}\label{sec:error}

% Since~\algname involves multiple components working jointly to synthesize trajectories for various manipulation tasks, herein we analyze the errors resulting from each component and how the overall system can be further improved.
In this section, we analyze the errors resulting from each component of~\algname and how the overall system can be further improved.
We conduct experiments in simulation where we have access to ground-truth perception and dynamics model (i.e., the simulator).
% Results are shown in Fig.~\ref{fig:error}
.
% ``Specification error'' here refers to the error by U-Net, such as noisy predictions that are difficult to be used by motion planners.
% For this baseline and~\algname, ``specification error'' refers to errors by LLMs either in composing policy logic or composing value maps.
% when composing policy logic, where common errors include incorrect specification of primitive parameters by LLMs.
``Dynamics error'' refers to errors made by the dynamics model\footnote{\textbf{LLM~+~Primitives}~\cite{liang2022code} does not use model-based planning, thus not having a dynamics module.}.
``Perception error'' refers to errors made by the perception module\footnote{\textbf{U-Net~+~MP}~\cite{sharma2022correcting} maps RGB-D to costmaps using U-Net~\cite{ronneberger2015u,cciccek20163d}, thus not having perception module. Errors by which are attributed to ``specification error''.}.
``Specification error'' refers to errors made by the module specifying cost or parameters for the low-level motion planner or primitives. Examples for each method include 1) noisy prediction by the U-Net, 2) incorrect parameters specified by the LLM, and 3) incorrect value maps specified by the LLM.
% In comparison, although~\algname uses multiple components, by formulating it as a joint model-based optimization problem,~\algname achieves the lowest overall error, and its biggest source of error is the perception module.
As shown in Fig.~\ref{fig:error},~\algname achieves lowest ``specification error'' due to its generalization and flexibility.
We also find that having access to a more robust perception pipeline and a physically-realistic dynamics model can contribute to better overall performance.
% We also observe that having access to a better dynamics model (rather than a heuristics-based model) can contribute to better overall performance, such as a learned model or a physics-based model.
This observation aligns with our real-world experiment, where most errors are from perception. For example, we find that the detector is sensitive to initial poses of objects and is less robust when detecting object parts.
% The tracker may also confuse two similar-looking objects (e.g., two drawer handles) under long-lasting and heavy occlusions.
% Depending on the specific object and its pose, the detector sometimes may output incorrect bounding boxes for the object. The errors are more pronounced when trying to detect object parts, such as a drawer handle. Another common perception errors are from the tracker. Sometimes it may confuse two similar-looking objects (e.g., two drawer handles) under long-lasting and heavy occlusions.}

\section{Conclusion, Limitations, \& Future Works}

In this work, we present\textbf{\textsc{~\algname}}, a general framework for extracting affordances and constraints, grounded in 3D perceptual space, from LLMs and VLMs for everyday manipulation tasks in the real world, offering significant generalization advantages for open-set instructions and objects.
% In particular, we use code-writing LLMs to interact with VLMs to compose 3D value maps grounded in observation space, which are used to synthesize trajectories for everyday manipulation tasks. Furthermore, we show \algname can benefit from online interactions by efficiently learning a dynamics model for contact-rich tasks.
Despite compelling results,~\algname has several limitations. First, it relies on external perception modules, which is limiting in tasks that require holistic visual reasoning or understanding of fine-grained object geometries.
Second, while applicable to efficient dynamics learning, a general-purpose dynamics model is still required to achieve contact-rich tasks with the same level of generalization.
Third, our motion planner considers only end-effector trajectories while whole-arm planning is also feasible and likely a better design choice~\cite{kavraki1996probabilistic,ratliff2018riemannian,marcucci2022motion}.
Finally, manual prompt engineering is required for LLMs.
We also see several exciting venues for future work. For instance, recent success of multi-modal LLMs~\cite{driess2023palm,openai2023gpt,li2023blip} can be directly translated into~\algname for direct visual grounding.
Methods developed for alignment~\cite{ouyang2022training,bai2022constitutional} and prompting~\cite{wei2022chain,wang2022self,kojima2022large,yao2023tree} may be used to alleviate prompt engineering effort.
% Finally, while we use greedy search in trajectory optimization, more advanced optimization methods can be developed that best interface with value maps synthesized by~\algname.
Finally, more advanced trajectory optimization methods can be developed that best interface with value maps synthesized by~\algname.

\label{sec:conclusion}

\acknowledgments{We would like to thank Andy Zeng, Igor Mordatch, and the members of the Stanford Vision and Learning Lab for the fruitful discussions. This work was in part supported by AFOSR YIP FA9550-23-1-0127, ONR MURI N00014-22-1-2740, ONR MURI N00014-21-1-2801, ONR N00014-23-1-2355, the Stanford Institute for Human-Centered AI (HAI), JPMC, and Analog Devices. Wenlong Huang is partially supported by Stanford School of Engineering Fellowship. Ruohan Zhang is partially supported by Wu Tsai Human Performance Alliance Fellowship.}

\bibliography{main}

\appendix
\newpage

\section{Appendix}

\subsection{Code Release}
We provide an open-sourced implementation of VoxPoser at~\href{https://github.com/huangwl18/VoxPoser}{github.com/huangwl18/VoxPoser} based on RLBench~\cite{james2020rlbench}, as its diversity of tasks and scenes best resembles our real-world setup.

\subsection{Emergent Behavioral Capabilities}

\begin{figure}[h]
  \begin{center}
  % \vspace{-1.8em}
    \includegraphics[width=0.85\linewidth]{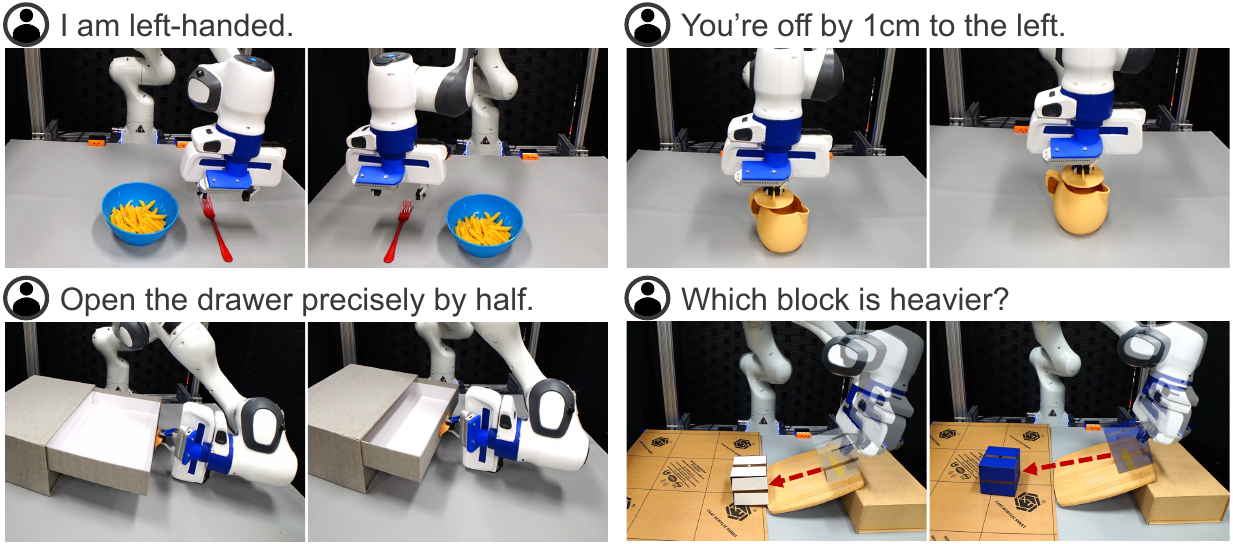}
    % \vspace{-1.4em}
    \caption{\small Emergent behavioral capabilities by~\algname inherited from the language model, including behavioral commonsense reasoning (\textbf{top left}), fine-grained language correction (\textbf{top right}), multi-step visual program (\textbf{bottom left}), and estimating physical properties of objects (\textbf{bottom right}). }
    \label{fig:emergent}
    % \vspace{-1em}
  \end{center}
\end{figure}

Emergent capabilities refer to unpredictable phenomenons that are only present in large models~\cite{wei2022emergent}. As~\algname uses pre-trained LLMs as backbone, we observe similar embodied emergent capabilities driven by the rich world knowledge of LLMs. In particular, we focus our study on the \emph{behavioral capabilities} that are unique to~\algname.
We observe the following capabilities:
\begin{itemize}
    \item \textbf{Behavioral Commonsense Reasoning}: During a task where robot is setting the table, the user can specify behavioral preferences such as ``I am left-handed'', which requires the robot to comprehend its meaning in the context of the task.~\algname decides that it should move the fork from the right side of the bowl to the left side.
    \item \textbf{Fine-grained Language Correction}: For tasks that require high precision such as ``covering the teapot with the lid'', the user can give precise instructions to the robot such as ``you're off by 1cm''.~\algname similarly adjusts its action based on the feedback.
    \item \textbf{Multi-step Visual Program}~\cite{gupta2023visual,suris2023vipergpt}: Given a task ``open the drawer precisely by half'' where there is insufficient information because object models are not available,~\algname can come up with multi-step manipulation strategies based on visual feedback that first opens the drawer fully while recording handle displacement, then close it back to the mid-point to satisfy the requirement.
    \item \textbf{Estimating Physical Properties}:
    Given two blocks of unknown mass, the robot is tasked to conduct physics experiments using an existing ramp to determine which block is heavier.~\algname decides to push both blocks off the ramp and choose the block traveling the farthest as the heavier block.
    Interestingly, this mirrors a common human oversight: in an ideal, frictionless world, both blocks would traverse the same distance under the influence of gravity. This serves as a lighthearted example that language models can exhibit limitations similar to human reasoning.
\end{itemize}

\newpage
\subsection{APIs for \algname}\label{app:api}

Central to~\algname is an LLM generating Python code that is executed by a Python interpreter. Besides exposing NumPy~\cite{harris2020array} and the~\href{http://matthew-brett.github.io/transforms3d/}{Transforms3d} library to the LLM, we provide the following environment APIs that LLMs can choose to invoke:

\texttt{\textbf{detect(obj\_name)}}: Takes in an object name and returns a list of dictionaries, where each dictionary corresponds to one instance of the matching object, containing center position, occupancy grid, and mean normal vector.

\texttt{\textbf{execute(movable,affordance\_map,avoidance\_map,rotation\_map,velocity\_map,gripper\_map)}}: Takes in an ``entity of interest'' as ``movable'' (a dictionary returned by \texttt{detect}) and (optionally) a list of value maps and invokes the motion planner to execute the trajectory. Note that in MPC settings, ``movable'' and the input value maps are functions that can be re-evaluated to reflect the latest environment observation.

\texttt{\textbf{cm2index(cm,direction)}}: Takes in a desired offset distance in centimeters along direction and returns 3-dim vector reflecting displacement in voxel coordinates.

\texttt{\textbf{index2cm(index,direction)}}: Inverse of \texttt{cm2index}. Takes in an integer ``index'' and a ``direction'' vector and returns the distance in centimeters in world coordinates displaced by the ``integer'' in voxel coordinates.

\texttt{\textbf{pointat2quat(vector)}}: Takes in a desired pointing direction for the end-effector and returns a satisfying target quaternion.

\texttt{\textbf{set\_voxel\_by\_radius(voxel\_map,voxel\_xyz,radius\_cm,value)}}: Assigns ``value'' to voxels within ``radious\_cm'' from ``voxel\_xyz'' in ``voxel\_map''.

\texttt{\textbf{get\_empty\_affordance\_map()}}: Returns a default affordance map initialized with 0, where a high value attracts the entity.

\texttt{\textbf{get\_empty\_avoidance\_map()}}: Returns a default avoidance map initialized with 0, where a high value repulses the entity.

\texttt{\textbf{get\_empty\_rotation\_map()}}: Returns a default rotation map initialized with current end-effector quaternion.

\texttt{\textbf{get\_empty\_gripper\_map()}}: Returns a default gripper map initialized with current gripper action, where 1 indicates ``closed'' and 0 indicates ``open''.

\texttt{\textbf{get\_empty\_velocity\_map()}}: Returns a default affordance map initialized with 1, where the number represents scale factor (e.g., 0.5 for half of the default velocity).

\texttt{\textbf{reset\_to\_default\_pose()}}: Reset to robot rest pose.

\newpage
\subsection{Real-World Environment Setup}\label{app:real}

We use a Franka Emika Panda robot with a tabletop setup. We use Operational Space Controller with impedance from Deoxys~\cite{zhu2022viola}.
We mount two RGB-D cameras (Azure Kinect) at two opposite ends of the table: bottom right and top left from the top down view. At the start of each rollout, both cameras start recording and return the real-time RGB-D observations at 20 Hz. 

For each task, we evaluate each method on two settings: without and with disturbances. For tasks with disturbances, we apply three kinds of disturbances to the environment, which we pre-select a sequence of them at the start of the evaluation: 1) random forces applied to the robot, 2) random displacement of task-relevant and distractor objects, and 3) reverting task progress (e.g., pull drawer open while it's being closed by the robot). We only apply the third disturbances to tasks where ``entity of interest'' is an object or object part.

We compare to a variant of Code as Policies~\cite{liang2022code} as a baseline that uses an LLM with action primitives. The primitives include: \texttt{move\_to\_pos}, \texttt{rotate\_by\_quat}, \texttt{set\_vel}, \texttt{open\_gripper}, \texttt{close\_gripper}. We do not provide primitives such as pick-and-place as they would be tailored for a particular suite of tasks that we do not constrain to in our study (similar to the control APIs for~\algname specified in Sec.~\ref{app:api}).

\subsubsection{Tasks}

\textbf{Move \& Avoid}: ``Move to the top of [obj1] while staying away from [obj2]'', where [obj1] and [obj2] are randomized everyday objects selected from the list: apple, banana, yellow bowl, headphones, mug, wood block.

\textbf{Set Up Table}: ``Please set up the table by placing utensils for my pasta''.

\textbf{Close Drawer}: ``Close the [deixis] drawer'', where [deixis] can be ``top'' or ``bottom''.

\textbf{Open Bottle}: ``Turn open the vitamin bottle''.

\textbf{Sweep Trash}: ``Please sweep the paper trash into the blue dustpan''.

\newpage
\subsection{Simulated Environment Setup}\label{app:sim}

We implement a tabletop manipulation environment with a Franka Emika Panda robot in SAPIEN~\cite{xiang2020sapien}. The controller takes as input a desired end-effector 6-DoF pose, calculates a sequence of interpolated waypoints using inverse kinematics, and finally follows the waypoints using a PD controller. We use a set of 10 colored blocks and 10 colored lines in addition to an articulated cabinet with 3 drawers. They are initialized differently depending on the specific task. The lines are used as visual landmarks and are not interactable. For perception, a total of 4 RGB-D cameras are mounted at each end of the table pointing at the center of the workspace.

\subsubsection{Tasks}

We create a custom suite of 13 tasks shown in Table~\ref{tab:sim_full}. Each task comes with a templated instruction (shown in Table~\ref{tab:sim_full}) where there may be one or multiple attributes randomized from the pre-defined list below. At reset time, a number of objects are selected (depending on the specific task) and are randomized across the workspace while making sure that task is not completed at reset and that task completion is feasible. A complete list of attributes can be found below, divided into ``seen'' and ``unseen'' categories:

\textbf{Seen Attributes: }

\begin{itemize}

\item \texttt{\textbf{[pos]}}: [``back left corner of the table'', ``front right corner of the table'', ``right side of the table'', ``back side of the table'']

\item \texttt{\textbf{[obj]}}: [``blue block'', ``green block'', ``yellow block'', ``pink block'', ``brown block'']

\item \texttt{\textbf{[preposition]}}: [``left of'', ``front side of'', ``top of'']

\item \texttt{\textbf{[deixis]}}: [``topmost'', ``second to the bottom'']

\item \texttt{\textbf{[dist]}}: [3, 5, 7, 9, 11]

\item \texttt{\textbf{[region]}}: [``right side of the table'', ``back side of the table'']

\item \texttt{\textbf{[velocity]}}: [``faster speed'', ``a quarter of the speed'']

\item \texttt{\textbf{[line]}}: [``blue line'', ``green line'', ``yellow line'', ``pink line'', ``brown line'']

\end{itemize}

\textbf{Unseen Attributes: }

\begin{itemize}

\item \texttt{\textbf{[pos]}}: [``back right corner of the table'', ``front left corner of the table'', ``left side of the table'', ``front side of the table'']

\item \texttt{\textbf{[obj]}}: [``red block'', ``orange block'', ``purple block'', ``cyan block'', ``gray block'']

\item \texttt{\textbf{[preposition]}}: [``right of'', ``back side of'']

\item \texttt{\textbf{[deixis]}}: [``bottommost'', ``second to the top'']

\item \texttt{\textbf{[dist]}}: [4, 6, 8, 10]

\item \texttt{\textbf{[region]}}: [``left side of the table'', ``front side of the table'']

\item \texttt{\textbf{[velocity]}}: [``slower speed'', ``3x speed'']

\item \texttt{\textbf{[line]}}: [``red line'', ``orange line'', ``purple line'', ``cyan line'', ``gray line'']

\end{itemize}

\newpage
\subsubsection{Full Results on Simulated Environments}
\begin{table}[H]
\footnotesize
\begin{center}
\setlength\tabcolsep{3pt}
\bgroup
\def\arraystretch{1.4}
\begin{tabular}{@{}lcccccc@{}}
    \toprule
      \multicolumn{1}{l}{} & \multicolumn{2}{c}{\textbf{U-Net + MP}} & \multicolumn{2}{c}{\textbf{LLM + Prim.}} &  \multicolumn{2}{c}{\textbf{\algname}} \\
      \cmidrule(lr){2-3} \cmidrule(lr){4-5} \cmidrule(lr){6-7}
    Tasks & SA & UA & SA & UA & SA & UA  \\
    \midrule
    \rowcolor{lightblue}
    \scriptsize{move to the [preposition] the [obj] }& 95.0\% &  0.0\% &   85.0\% &  60.0\% &  90.0\% &  55.0\%  \\
    \rowcolor{lightblue}
    \scriptsize{move to the [pos] while staying on the [preposition] the [obj]}& 100.0\% & 10.0\% &  80.0\% &  30.0\% &  95.0\% &  50.0\%  \\
    \rowcolor{lightblue}
    \scriptsize{move to the [pos] while moving at [velocity] when within [dist]cm from the {obj}}& 80.0\% &  0.0\% &   10.0\% &  0.0\% &   100.0\% & 95.0\%  \\
    \rowcolor{lightblue}
    \scriptsize{close the [deixis] drawer by pushing} & 0.0\% &   0.0\% &   60.0\% &  60.0\% &  80.0\% &  80.0\%  \\
    \rowcolor{lightblue}
    \scriptsize{push the [obj] along the [line]} & 0.0\% &   0.0\% &   0.0\% &   0.0\% &   65.0\% &  30.0\%  \\
    \rowcolor{lightblue}
    \scriptsize{grasp the [obj] from the table at [velocity]} & 35.0\% &  0.0\% &   75.0\% &  70.0\% &  65.0\% &  40.0\%  \\
    \rowcolor{lightblue}
    \scriptsize{drop the [obj] to the [pos]} & 70.0\% &  10.0\% &  60.0\% &  100.0\% & 60.0\% &  100.0\%  \\
    \rowcolor{lightblue}
    \scriptsize{push the [obj] while letting it stay on [region]} & 0.0\% &   5.0\% &   10.0\% &  0.0\% &   50.0\% &  50.0\%  \\
    \rowcolor{lightorange}
    \scriptsize{move to the [region]} & 5.0\% &   0.0\% &   100.0\% & 95.0\% &  100.0\% & 100.0\%  \\
    \rowcolor{lightorange}
    \scriptsize{move to the [pos] while staying at least [dist]cm from the [obj]} & 0.0\% &   0.0\% &   15.0\% &  20.0\% &  85.0\% &  90.0\%  \\
    \rowcolor{lightorange}
    \scriptsize{move to the [pos] while moving at [velocity] in the [region]} & 0.0\% &   0.0\% &   90.0\% &  45.0\% &  85.0\% &  85.0\%  \\
    \rowcolor{lightorange}
    \scriptsize{push the [obj] to the [pos] while staying away from [obstacle]} & 0.0\% &   0.0\% &   0.0\% &   10.0\% &  45.0\% &  55.0\%  \\
    \rowcolor{lightorange}
    \scriptsize{push the [obj] to the [pos]} & 0.0\% &   0.0\% &   20.0\% &  25.0\% &  80.0\% &  75.0\% \\
    \bottomrule
\end{tabular}
\egroup
\caption{\small Full experimental results in simulation on \textcolor{darkblue}{seen tasks} and \textcolor{darkorange}{unseen tasks}. ``SA'' indicates seen attributes and ``UA'' indicates unseen attributes. Each entry represents success rate averaged across 20 episodes.}
\label{tab:sim_full}
\end{center}
\end{table}

\newpage
\subsection{Prompts}\label{app:prompts}

Prompts used in Sec.~\ref{sec:real} and Sec.~\ref{sec:sim} can be found below. 

\texttt{\textbf{planner}}: Takes in a user instruction $\mathcal{L}$ and generates a sequence of sub-tasks $\ell_i$ which is fed into ``composer'' (Note that planner is not used in simulation as the evaluated tasks consist of a single manipulation phase). 

real-world: \href{https://voxposer.github.io/prompts/real_planner_prompt.txt}{voxposer.github.io/prompts/real\_planner\_prompt.txt}.

\texttt{\textbf{composer}}: Takes in sub-task instruction $\ell_i$ and invokes necessary value map LMPs to compose affordance maps and constraint maps. 

simulation: \href{https://voxposer.github.io/prompts/sim_composer_prompt.txt}{voxposer.github.io/prompts/sim\_composer\_prompt.txt}.

real-world: \href{https://voxposer.github.io/prompts/real_composer_prompt.txt}{voxposer.github.io/prompts/real\_composer\_prompt.txt}.

\texttt{\textbf{parse\_query\_obj}}: Takes in a text query of object/part name and returns a list of dictionaries, where each dictionary corresponds to one instance of the matching object containing center position, occupancy grid, and mean normal vector. 

simulation: \href{https://voxposer.github.io/prompts/sim_parse_query_obj_prompt.txt}{voxposer.github.io/prompts/sim\_parse\_query\_obj\_prompt.txt}.

real-world: \href{https://voxposer.github.io/prompts/real_parse_query_obj_prompt.txt}{voxposer.github.io/prompts/real\_parse\_query\_obj\_prompt.txt}.

\texttt{\textbf{get\_affordance\_map}}: Takes in natural language parametrization from \texttt{composer} and returns a NumPy array for task affordance map. 

simulation: \href{https://voxposer.github.io/prompts/sim_get_affordance_map_prompt.txt}{voxposer.github.io/prompts/sim\_get\_affordance\_map\_prompt.txt}.

real-world: \href{https://voxposer.github.io/prompts/real_get_affordance_map_prompt.txt}{voxposer.github.io/prompts/real\_get\_affordance\_map\_prompt.txt}.

\texttt{\textbf{get\_avoidance\_map}}: Takes in natural language parametrization from \texttt{composer} and returns a NumPy array for task avoidance map. 

simulation: \href{https://voxposer.github.io/prompts/sim_get_avoidance_map_prompt.txt}{voxposer.github.io/prompts/sim\_get\_avoidance\_map\_prompt.txt}.

real-world: \href{https://voxposer.github.io/prompts/real_get_avoidance_map_prompt.txt}{voxposer.github.io/prompts/real\_get\_avoidance\_map\_prompt.txt}.

\texttt{\textbf{get\_rotation\_map}}: Takes in natural language parametrization from \texttt{composer} and returns a NumPy array for end-effector rotation map. 

simulation: \href{https://voxposer.github.io/prompts/sim_get_rotation_map_prompt.txt}{voxposer.github.io/prompts/sim\_get\_rotation\_map\_prompt.txt}.

real-world: \href{https://voxposer.github.io/prompts/real_get_rotation_map_prompt.txt}{voxposer.github.io/prompts/real\_get\_rotation\_map\_prompt.txt}.

\texttt{\textbf{get\_gripper\_map}}: Takes in natural language parametrization from \texttt{composer} and returns a NumPy array for gripper action map. 

simulation: \href{https://voxposer.github.io/prompts/sim_get_gripper_map_prompt.txt}{voxposer.github.io/prompts/sim\_get\_gripper\_map\_prompt.txt}.

real-world: \href{https://voxposer.github.io/prompts/real_get_gripper_map_prompt.txt}{voxposer.github.io/prompts/real\_get\_gripper\_map\_prompt.txt}.

\texttt{\textbf{get\_velocity\_map}}: Takes in natural language parametrization from \texttt{composer} and returns a NumPy array for end-effector velocity map. 

simulation: \href{https://voxposer.github.io/prompts/sim_get_velocity_map_prompt.txt}{voxposer.github.io/prompts/sim\_get\_velocity\_map\_prompt.txt}.

real-world: \href{https://voxposer.github.io/prompts/real_get_velocity_map_prompt.txt}{voxposer.github.io/prompts/real\_get\_velocity\_map\_prompt.txt}.

\end{document}